\documentclass[sigconf]{acmart}

\usepackage{booktabs} 
\usepackage[utf8]{inputenc} 
\usepackage[T1]{fontenc}    
\usepackage[ruled,vlined,linesnumbered]{algorithm2e}
\usepackage{subfigure}
\usepackage{graphicx}

\DeclareMathOperator*{\argmin}{arg\,min}

\setcopyright{acmlicensed}

\acmDOI{10.1145/3605098.3636113}

\acmISBN{979-8-4007-0243-3/24/04}

\acmConference[SAC'24]{ACM SAC Conference}{April 8 –April 12, 2024}{Avila, Spain}
\acmYear{2024}
\copyrightyear{2024}

\acmArticle{4}
\acmPrice{15.00}

\begin{document}
\title{Bayesian Soft Actor-Critic: A Directed Acyclic Strategy Graph Based Deep Reinforcement Learning}
  
\renewcommand{\shorttitle}{SIG Proceedings Paper in LaTeX Format}

\author{Qin Yang}
\authornote{Corresponding author: Dr. Yang; Email: RickYang2014@gmail.com}
\affiliation{%
  \institution{Computer Science and Information Systems Department, Braldey University}
  \streetaddress{1501 W Bradley Ave, Bradley Hall 195}
  \city{Peoria}
  \state{Illinois}
  \country{U.S.}
  \postcode{61625}
}

\author{Ramviyas Parasuraman}
\affiliation{%
  \institution{Department of Computer Science, University of Georgia}
  \streetaddress{415 Boyd Research and Education Center
University of Georgia}
  \city{Athens} 
  \state{Georgia}
  \country{U.S.}
  \postcode{30602-7404}
}





\begin{abstract}
Adopting reasonable strategies is challenging but crucial for an intelligent agent with limited resources working in hazardous, unstructured, and dynamic environments to improve the system's utility, decrease the overall cost, and increase mission success probability.
This paper proposes a novel directed acyclic strategy graph decomposition approach based on Bayesian chaining to separate an intricate policy into several simple sub-policies and organize their relationships as Bayesian strategy networks (BSN). We integrate this approach into the state-of-the-art DRL method -- soft actor-critic (SAC), and build the corresponding Bayesian soft actor-critic (BSAC) model by organizing several sub-policies as a joint policy. We compare our method against the state-of-the-art deep reinforcement learning algorithms on the standard continuous control benchmarks in the OpenAI Gym environment. The results demonstrate that the promising potential of the BSAC method significantly improves training efficiency.
\end{abstract}

%
%



\maketitle

\section{Introduction}

A strategy is a rule used by agents to select an action to pursue goals, which is equivalent to a policy in a Markov Decision Process (MDP) \cite{rizk2018decision}. It exhibits the fundamental properties of agents' perception, reasoning, planning, decision-making, learning, problem-solving, and communication in interaction with dynamic and complex environments \cite{langley2009cognitive,yang2019self}. Especially in the field of real-time strategy (RTS) game \cite{buro2003real,yang2022game,yang2023hierarchical} and real-world implementation scenarios like robot-aided urban search and rescue (USAR) missions \cite{murphy2014disaster,yang2020needs}, agents need to dynamically change the strategies adapting to the current situations based on the environments and their expected utilities or needs \cite{yang2020hierarchical, yang2021can}.

\begin{figure}[t]
\centering
\includegraphics[width=1\columnwidth]{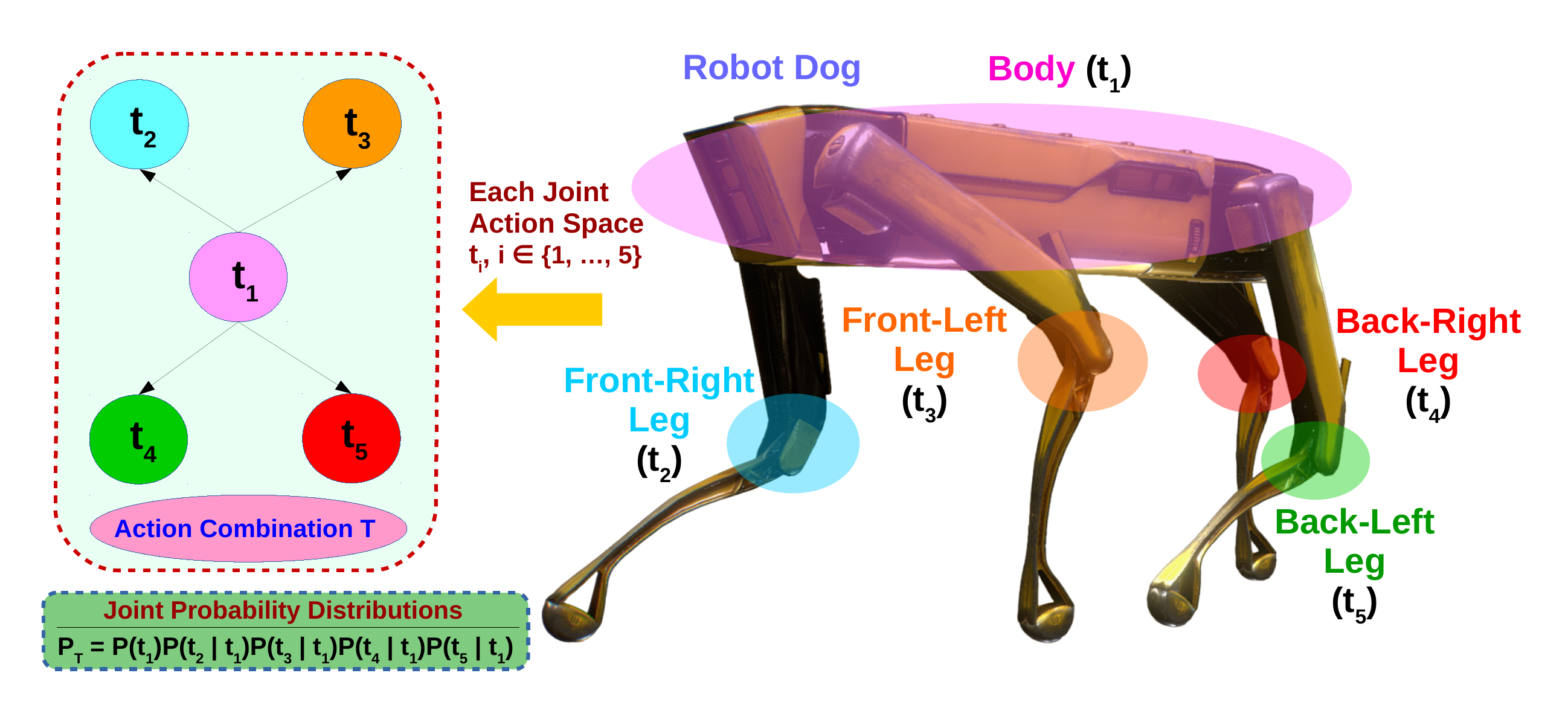}
\caption{An example of the robot dog model's Bayesian Strategy Network, showing a decomposed strategy based on action dependencies.}
\label{fig:overview}
\end{figure}


\begin{figure*}[tbp]
\centering
\includegraphics[width=2\columnwidth]{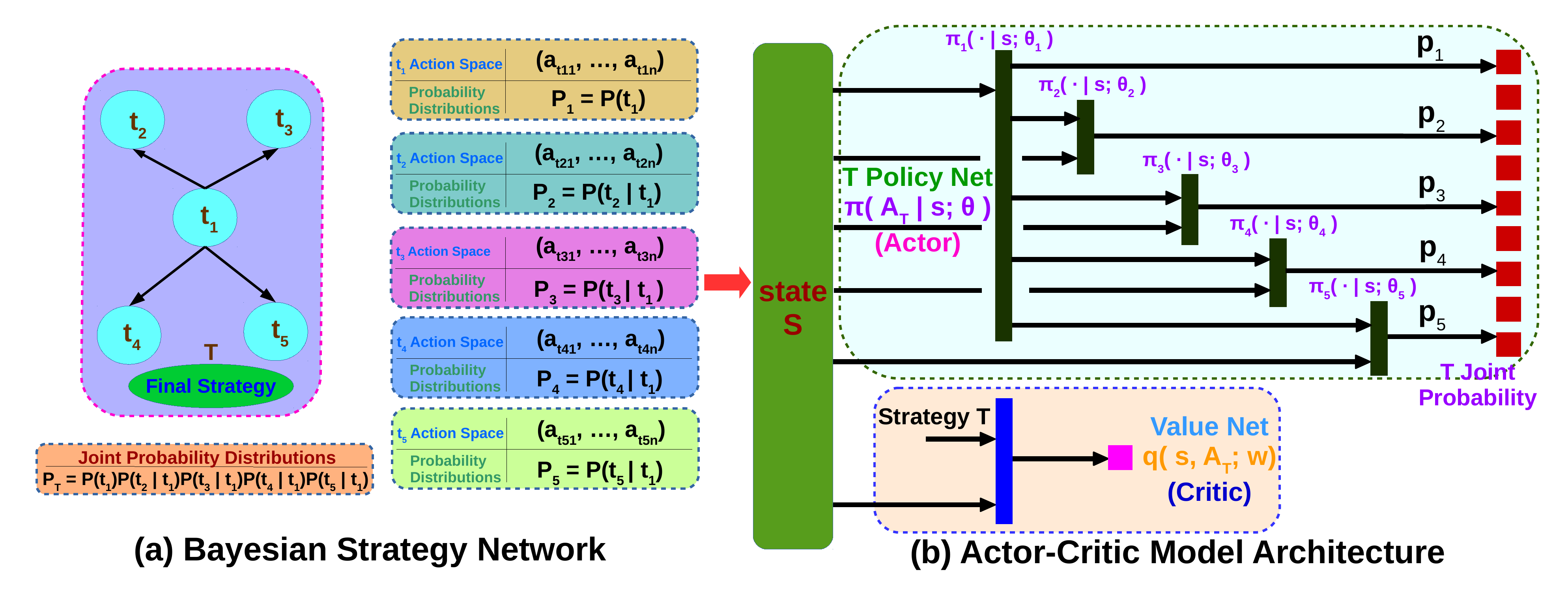}
\caption{An example of the proposed BSN based implementation of the Actor-Critic DRL architecture model.}
\label{bsn_drl}
\end{figure*}

Furthermore, in task-oriented decision-making, hierarchical reinforcement learning (HRL) enables autonomous decomposition of challenging long-horizon decision-making tasks into simpler subtasks \cite{pateria2021hierarchical}. 
However, a single strategy might involve learning several policies simultaneously, which means the strategy consists of several tactics (sub-strategies) or actions executing a simple task, especially in the robot locomotion \cite{polydoros2017survey} and RTS game \cite{shao2019survey} domain. 
Recently, the soft actor-critic (SAC) approach \cite{haarnoja2018soft}, an off-policy actor-critic algorithm based on the maximum entropy framework, has shown to be one of the leading approaches for model-free off-policy DRL and one of the promising algorithms implemented in the real robot domain \cite{haarnoja2018soft2}.
But, DRL is still hard to explain formally how and why the randomization works, which brings the difficulty of designing efficient models expressing the relationships between various strategies (policies) \cite{zhao2020sim}. 

To address this gap, this paper first introduces the Bayesian Strategy Network (BSN) based on the Bayesian net to decompose a complex strategy or intricate behavior into several simple tactics or actions. An example of a BSN-based strategy decomposition (or action dependencies) of a robot dog is shown in Fig.~\ref{fig:overview}. Furthermore, we propose a new DRL model termed Bayesian Soft Actor-Critic (BSAC), which integrates the Bayesian Strategy Networks (BSN) and the state-of-the-art SAC method. 
We demonstrate the effectiveness of the BSAC against the other state-of-the-art approaches on the standard continuous control benchmark domains in the OpenAI Gym environment. 

\section{Methodology}
\label{bsac_method}

Building on top of the well-established suite of actor-critic methods, we introduce the Bayesian Strategy Network (BSN) and integrate the idea of the maximum entropy reinforcement learning framework in SAC, designing the method that results in our Bayesian soft actor-critic (BSAC) approach.

\subsection{Bayesian Strategy Networks (BSN)}
\label{section_bsn}
Supposing that the strategy $\mathcal{T}$ consists of $m$ tactics ($t_1, \dots, t_m$) and their specific relationships can be described as the BSN. We consider the probability distribution $P_{i}$ as the policy for tactic $t_i$. Then, according to the Bayesian chain rule, the joint policy $\pi(a_{\mathcal{T}} \in {\mathcal{T}},s)$ can be described as the joint probability function (Eq. \eqref{agent_task_policy}) through each sub-policy $\pi_i(t_i,s)$, correspondingly. 
An overview of the example BSN implementation in actor-critic architecture is presented in Fig. \ref{bsn_drl}. 
\begin{equation}
\begin{split}
    \pi_{\mathcal{T}}(t_1, \dots, t_m) = \pi_{1}(t_1) \prod_{i=2}^{m} \pi_{i}(t_i | t_1, \dots, t_i),~~~m \in Z^+.
\label{agent_task_policy}
\end{split}
\end{equation}

\subsection{Derivation of Sub-Policy Iteration}
Considering that the agent interacts with an environment through a sequence of observations, strategy (action combinations), and rewards, we can describe the relationships between actions in the strategy as a BSN, represented in Eq. \eqref{agent_task_policy} accordingly. More formally, we can use the corresponding deep convolution neural networks to approximate the strategy \textit{Policy Network} (Actor) in the Eq. \eqref{agent_task_policy} as Eq. \eqref{joint_task_policy} \footnote{Here, $\mathcal{A}$ is the joint action or strategy space for the policy $\pi$; $\theta_i$ and $a_{i_t}$ are the parameters and action space of each sub-policy network $\pi_i$.}. The \textit{Value Network} (Critic) $q(s, \mathcal{A}_t; w)$ evaluate the performance of the specific joint action $\mathcal{A}$ using the value function in Eq. \eqref{joint_task_value} with a parameter $w$. Then, we can calculate the corresponding parameters' gradient descent using Eq. \eqref{gradient_joint_task_policy}.
Through this process, we decompose the strategy policy network $\pi_{\mathcal{T}}$ into several sub-policies networks $\pi_i$ and organize them as the corresponding BSN. Furthermore, according to Eq. \eqref{gradient_joint_task_policy}, each sub-policies uses the same value network to update its parameters in every iteration.
\begin{flalign}
    & \pi(\mathcal{A}_t, s) \approx \pi(\mathcal{A}_t | s; \theta) = \prod_{i=1}^{m} \pi_i(a_{i_t} | s; \theta_i)
\label{joint_task_policy} \\
    & V(s; \theta, w) = \sum_{t \in T}\pi(\mathcal{A}_t | s; \theta) \cdot q(s, \mathcal{A}_t; w)
\label{joint_task_value} \\
    & \frac{\partial V(s; \theta, w)}{\partial \theta}
     = \sum_{i=1}^m \mathop{\mathbb{E}} \left[ \frac{\partial \log \pi_i(a_{i_t} | s; \theta_i)}{\partial \theta} \cdot q(s, \mathcal{A}_t; w) \right]
\label{gradient_joint_task_policy}
\end{flalign}

\subsection{Bayesian Soft Actor-Critic (BSAC)}

Our method incorporates the maximum entropy concept into the actor-critic deep RL algorithm. According to the definition of conditional entropy \cite{wehrl1978general}, joint entropy is equal to the sum of the conditional entropies of the variables in the set, which also follows the chain rule in probability theory.For each step of soft policy iteration, the joint policy $\pi$ will calculate the value to maximize the sum of sub-systems' $\pi_i$ entropy in the BSN using the below objective function Eq. \eqref{bsac_entropy}. In order to simplify the problem, we assume that the weight and the corresponding temperature parameters $\alpha_i$ for each action is the same in each sub-system.
\begin{equation}
\begin{split}
    J_V(\pi) = \sum_{t=0}^{T} \mathbb{E}_{(s_t, \mathcal{A}_t) \sim \rho_{\pi_i}} \left [r(s_t, \mathcal{A}_t) + \frac{\alpha}{m} \sum_{i=1}^{m} \mathcal{H}(\pi_i(\cdot|s_t)) \right ]
\label{bsac_entropy}
\end{split}
\end{equation}

The soft Q-value can be computed iteratively, starting from any function $Q: S \times A \rightarrow \mathbb{R}$ and repeatedly applying a modified Bellman backup operator $\mathcal{T}^\pi$ \cite{haarnoja2018soft}. In the Bayesian Soft Actor-Critic, the $\mathcal{T}^\pi$ is given by Eq. \eqref{bsac_bell}. Considering that the evaluation of each sub-policy applies the same Q-value and weight, the soft state value function can be represented in Eq. \eqref{bsac_sqv} \footnote{Here, $Z^{\pi_{old}}(s_t)$ is the partition function to normalize the distribution.}.
\begin{equation}
\begin{split}
    \mathcal{T}^\pi Q(s_t, \mathcal{A}_t) \triangleq r(s_t, \mathcal{A}_t) + \gamma \mathbb{E}_{s_{t+1} \sim p} \left [V(s_{t+1}) \right]
\label{bsac_bell}
\end{split}
\end{equation}
\begin{equation}
\begin{split}
    V(s_t) = \mathbb{E}_{\mathcal{A}_t \sim \pi_\phi} \left [Q(s_t, \mathcal{A}_t) - \frac{1}{m} \sum_{i=1}^{m} \log \pi_{\phi_i} (a_{i_t} | s_t)) \right]
\label{bsac_sqv}
\end{split}
\end{equation}

Specifically, in each sub-policy $\pi_i$ improvement step, for each state, we update the corresponding policy according to Eq. \eqref{bsac_pi}. 
\begin{equation}
\begin{split}
    \pi_{new} = \argmin_{\pi' \in \Pi} D_{KL} \left( \frac{1}{m} \prod_{i=1}^{m} \pi'_{i}(\cdot | s_t) \bigg\Vert \frac{exp(Q^{\pi_{old}}(s_t, \cdot))}{Z^{\pi_{old}}(s_t)} \right)
\label{bsac_pi}
\end{split}
\end{equation}


\begin{figure*}[tbp]
\centering
\subfigure[]{
     \label{fig:hopper}
     \includegraphics[width=0.55\columnwidth]{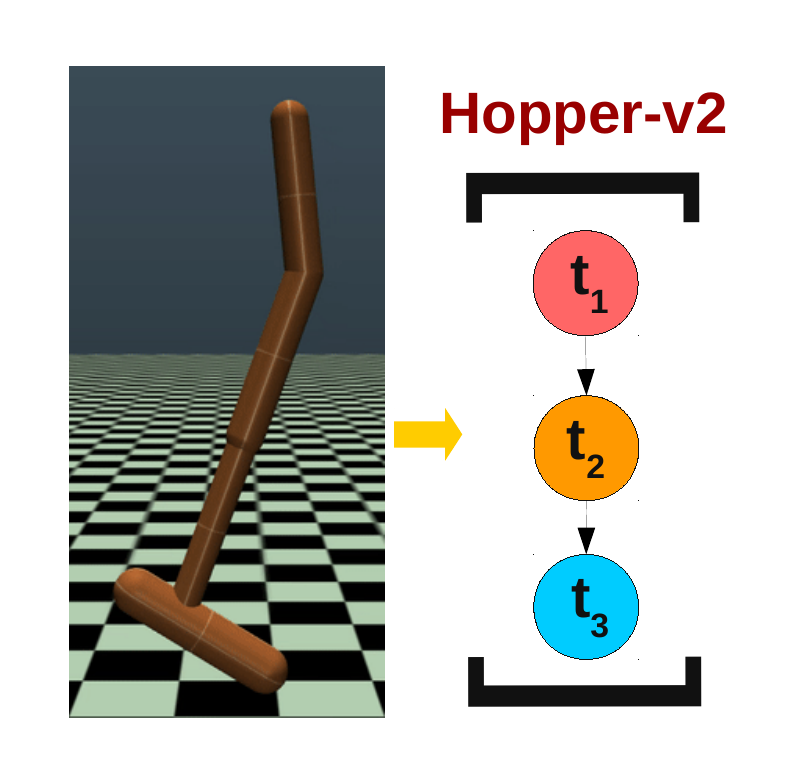}}
\subfigure[]{
     \includegraphics[width=0.65\columnwidth]{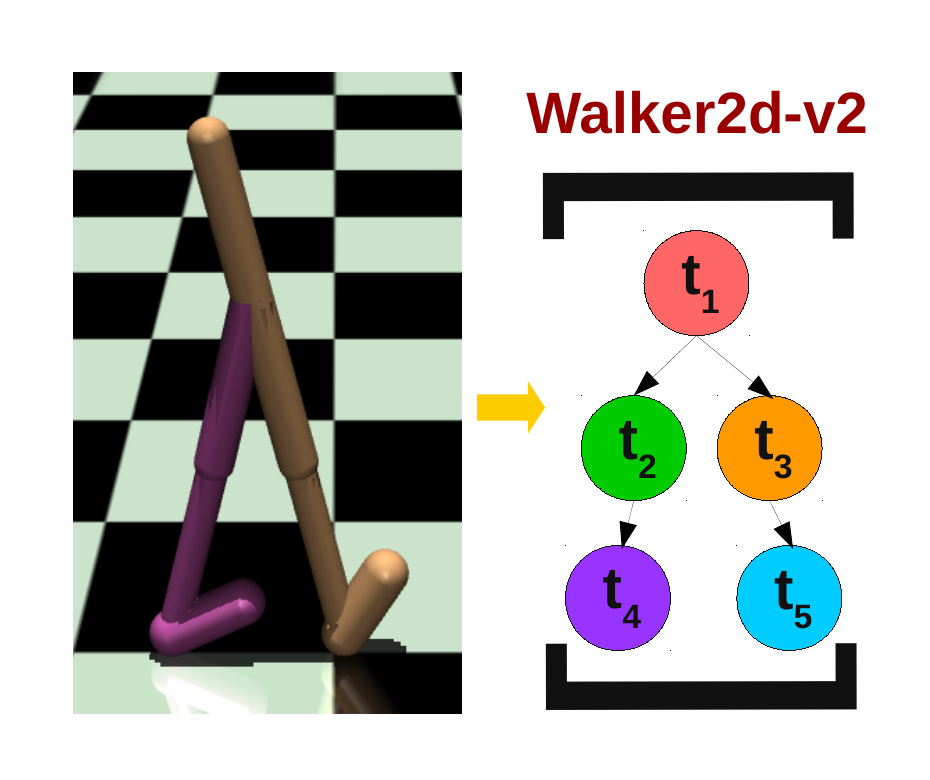}
     \label{fig:walker2d}}
\subfigure[]{
     \includegraphics[width=0.8\columnwidth]{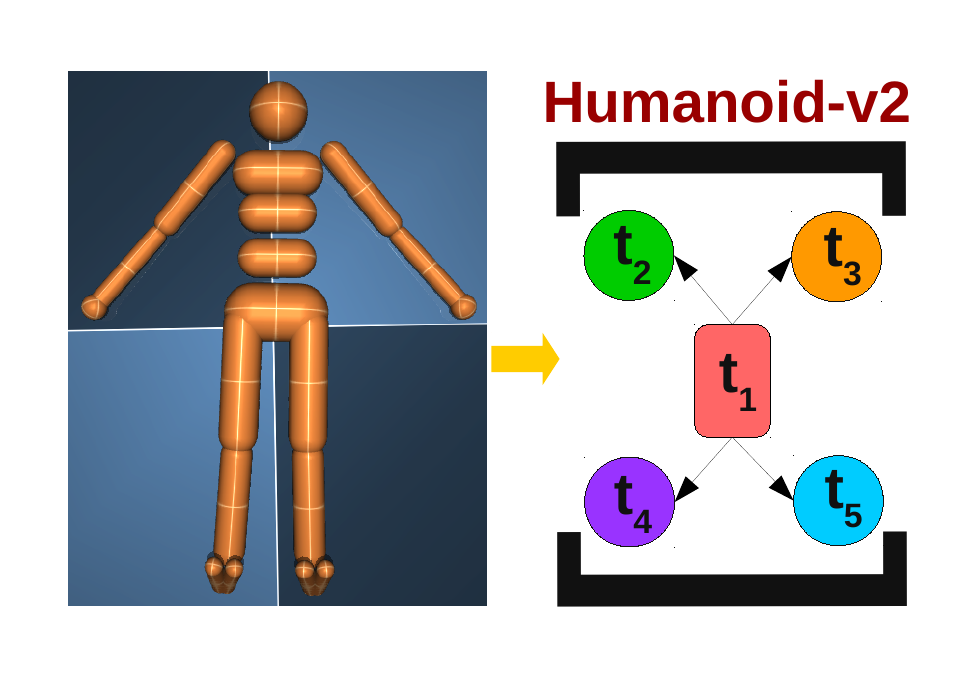}
     \label{fig:humanoid}}
\caption{The BSN model representing action dependencies on the Hopper-v2, Walker2d-v2 and Humanoid-v2 domains.}
\label{model_2v_bsn}
\end{figure*}
\begin{figure*}[tbp]
\centering
\subfigure[]{
\label{fig:hopper_r}
\includegraphics[width=0.325\textwidth]{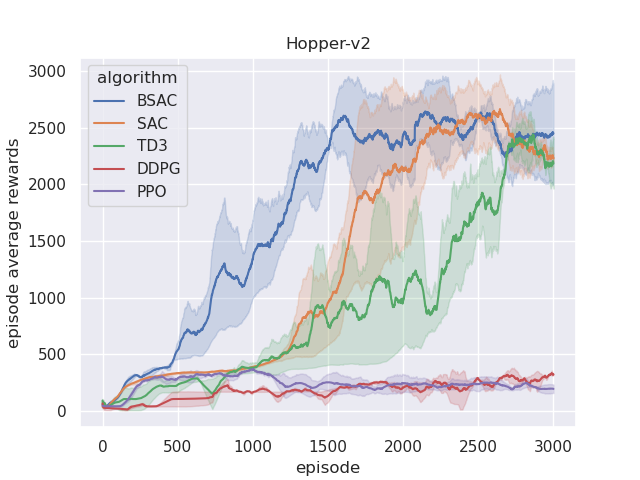}}
\subfigure[]{
\label{fig:walker2d_r}
\includegraphics[width=0.325\textwidth]{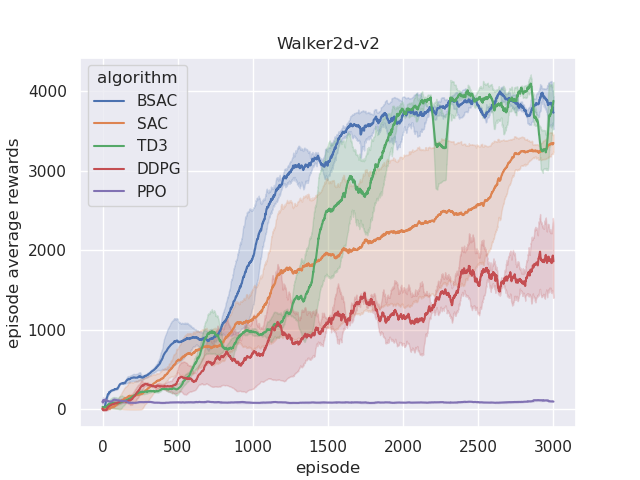}}
\subfigure[]{
\label{fig:humanoid_r}
\includegraphics[width=0.325\textwidth]{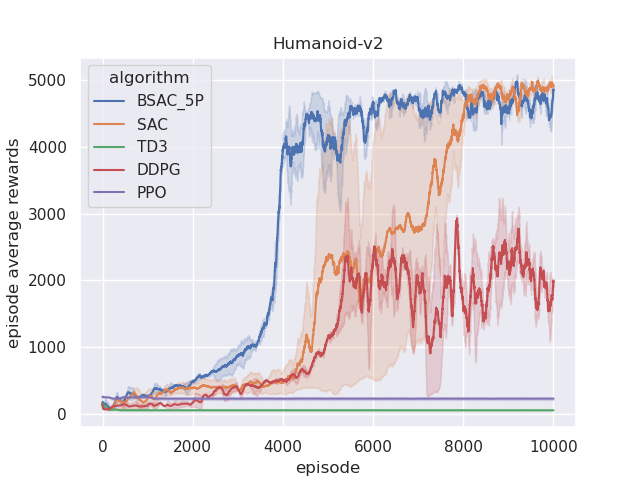}}
\caption{The performance of the standard continuous control benchmarks  comparison in the OpenAI Gym}
\label{bsn_gym}
\end{figure*}

\section{Experiments and Results}

In our experiments, we use three of the standard continuous control benchmark domains -- Hopper-v2, Walker2d-v2, and Humanoid-v2 in the OpenAI's Gym environment.
We study the performance of the proposed BSAC against the state-of-the-art continuous control algorithm, the SAC \cite{haarnoja2018soft} and other benchmark DRL algorithms, PPO \cite{schulman2017proximal}, DDPG \cite{lillicrap2015continuous}, and TD3 \cite{fujimoto2018addressing}.

In the corresponding experiments, we decompose hopper, walker2d, and humanoid behaviors into different BSNs. Such as three policies (3P) BSN (hip action $t_1$, knee action $t_2$, and ankle action $t_3$) in Hooper-v2 (Fig. \ref{fig:hopper}), five policies (5P) BSN (hip action $t_1$, left knee action $t_2$, right knee action $t_3$, left ankle action $t_4$, and right ankle action $t_5$) in Walker2d-v2 (Fig. \ref{fig:walker2d}), and five policies (5P) BSN (abdomen action $t_1$, the actions $t_2$ of right hip and right knee, the actions $t_3$ of left hip and left knee, the actions $t_4$ of right shoulder and right elbow, and actions $t_5$ of left shoulder and left elbow) in Humanoid-v2 (Fig.~\ref{fig:humanoid}). Furthermore, for corresponding BSAC models, we can formalize their joint policy (action) as Eq. \eqref{hopper_bsn}, Eq. \eqref{walker_bsn}, and Eq. \eqref{humanoid_bsn}.
\begin{equation}
\begin{split}
    P(t_1, t_2, t_3) = P(t_1) P(t_2|t_1) P(t_3|t_2)
\label{hopper_bsn}
\end{split}
\end{equation}
\vspace{-3mm}
\begin{equation}
\begin{split}
    P(t_1, t_2, t_3, t_4, t_5) = P(t_1)P(t_2|t_1)P(t_3|t_1)P(t_4|t_2)P(t_5|t_3)
\label{walker_bsn}
\end{split}
\end{equation}
\vspace{-3mm}
\begin{equation}
\begin{split}
    P(t_1, t_2, t_3, t_4, t_5) = P(t_1)P(t_2|t_1)P(t_3|t_1)P(t_4|t_1)P(t_5|t_1)
\label{humanoid_bsn}
\end{split}
\end{equation}

Comparing the performance of the BSAC with SAC, TD3, DDPG, and PPO in Hopper-v2 (Fig. \ref{fig:hopper_r}), Walker2d-v2 (Fig. \ref{fig:walker2d_r}), and Humanoid-v2 (Fig. \ref{fig:humanoid_r}), we prove that BSAC can achieve higher performance than other DRL algorithms. Furthermore, with the increasing complexity of the agent's behaviors and strategy, decomposing the complex behaviors into simple actions or tactics and organizing them as a suitable BSN, building the corresponding joint policy model in the BSAC can substantially increase training efficiency\footnote{We open the source of the BSAC algorithm on GitHub: \url{https://github.com/RickYang2016/Bayesian-Soft-Actor-Critic--BSAC}.}.

\begin{figure*}[tbp]
\centering
\begin{minipage}[b]{0.48\linewidth}
\includegraphics[width=0.8\columnwidth]{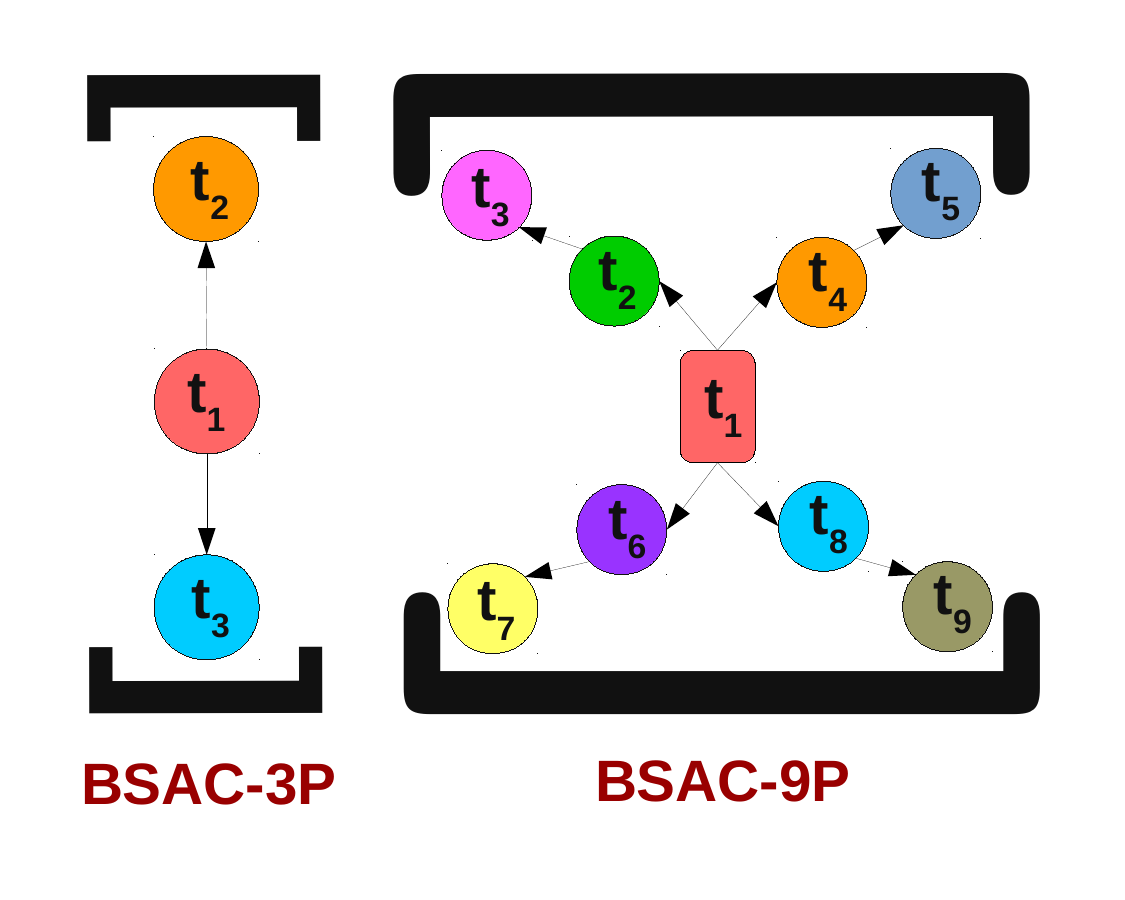}
\caption{Illustration of three policies (BSAC-3P) and nine policies (BSAC-9P) in the Humanoid-v2 domain}
\label{fig:humanoid_mb}
\end{minipage}
\hspace{3mm}
\begin{minipage}[b]{0.48\linewidth}
\includegraphics[width=0.9\textwidth]{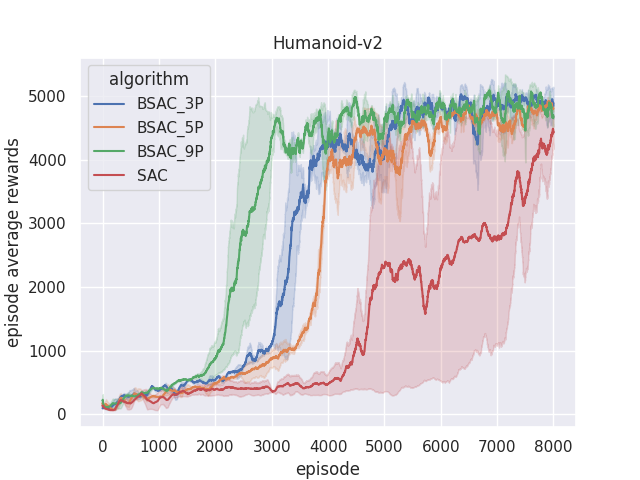}
\caption{The performance comparison of BSAC-3P, BSAC-5P, BSAC-9P, and SAC in the Humanoid-v2 domain}
\label{fig:humanoid_mr}
\end{minipage}
\end{figure*}

\paragraph{Comparing with Different BSN Models}
This section analyzes two other BSN models corresponding to different action decomposition methods in the Humanoid-v2 domain. In Fig. \ref{fig:humanoid_mb}, for the BSAC three sub-policies model (BSAC-3P), it generates the distribution of the abdomen actions $P(t_1)$, the distribution $P(t_2|t_1)$ of the actions of shoulder and elbow, and the distribution $P(t_3|t_1)$ of the actions of hip and knee, respectively. Within a sub-space, the actions are independent. For example, we do not consider the conditional dependence between the actions of the shoulder and elbow, but merging them as one joint action ($t_2$) depends on the abdomen joint ($t_1$). And the same situation is in the joints ($t_3$) of the hip and knee actions. The results shown in Fig. \ref{fig:humanoid} demonstrate that all BSAC models can achieve higher performance than the SAC. On the other hand, compared to the BSAC-3P models' performance, the BSAC-9P presents more advantages than BSAC-5P and BSAC-3P, and the five sub-policies model presents the worst performance among them. It implies that the joint policy distribution designed in the BSAC-9P model is more similar to the Q-value distribution than the other BSAC models and describes the more reasonable relationships among those actions in the current reward mechanism.


\section{Conclusions}
We introduce a novel directed acyclic strategy graph termed Bayesian Strategy Network (BSN) to achieve deep reinforcement learning (DRL) sample efficiency and improve training speed. 
Based on the Soft Actor-Critic (SAC) algorithm, we propose a new DRL model termed the Bayesian Soft Actor-Critic (BSAC), which integrates the BSN and forms a joint policy better adapting the Q-value distribution. Moreover, we demonstrate it on the standard continuous control benchmark and the results show the potential and significance of the proposed BSAC architecture.

\bibliographystyle{ACM-Reference-Format}
\bibliography{sample-bibliography} 

\end{document}